\documentclass[letterpaper, 10 pt, conference]{ieeeconf}  

\IEEEoverridecommandlockouts                              

\usepackage{cite}
\usepackage{amsmath,amssymb,amsfonts}
\usepackage{algorithmic}
\usepackage{arydshln}
\usepackage{graphicx}
\usepackage{textcomp}
\usepackage{xcolor}
\usepackage{booktabs} 
\usepackage[hidelinks]{hyperref}

\title{\LARGE \bf
Generating Heterogeneous Multi-dimensional Data : A Comparative Study
}

\author{Corbeau Michael$^{1}$, Claeys Emmanuelle$^{1}$, Serrurier Mathieu$^{1}$, Zaraté Pascale$^{1}$
\thanks{$^{1}$Institut de Recherche en Informatique de Toulouse,
        Cr Rose Dieng-Kuntz, 31400 Toulouse, France}
}

\begin{document}

\maketitle

\begin{abstract}

Allocation of personnel and material resources is highly sensible in the case of firefighter interventions. This allocation relies on simulations to experiment with various scenarios. The main objective of this allocation is the global optimization of the firefighters response. Data generation is then mandotory to study various scenarios

In this study, we propose to compare different data generation methods. Methods such as Random Sampling, Tabular Variational Autoencoders, standard Generative Adversarial Networks, Conditional Tabular Generative Adversarial Networks and Diffusion Probabilistic Models are examined to ascertain their efficacy in capturing the intricacies of firefighter interventions. 
Traditional evaluation metrics often fall short in capturing the nuanced requirements of synthetic datasets for real-world scenarios. To address this gap, an evaluation of synthetic data quality is conducted using a combination of domain-specific metrics tailored to the firefighting domain and standard measures such as the Wasserstein distance. Domain-specific metrics include response time distribution, spatial-temporal distribution of interventions, and accidents representation. These metrics are designed to assess data variability, the preservation of fine and complex correlations and anomalies such as event with a very low occurrence, the conformity with the initial statistical distribution and the operational relevance of the synthetic data. The distribution has the particularity of being highly unbalanced, none of the variables following a Gaussian distribution, adding complexity to the data generation process.

\end{abstract}

\section{Introduction}

In firefighting operations, optimizing resource allocation and deployment strategies is crucial for enhancing response efficiency and effectiveness. Simulations play a pivotal role in this optimization process, enabling the exploration of various scenarios and the development of robust strategies. However, these simulations necessitate high-quality synthetic data that accurately reflect the heterogeneity and imbalances inherent in real-world datasets.

Traditional data generation methods, often applied to benchmark datasets or used in the creation of text and images, fall short when tasked with producing synthetic data that mirror the complexities of real-world scenarios. This gap is particularly pronounced in the context of generating data for simulating firefighting interventions, where the data must encapsulate a wide range of variables, including geographical coordinates, incident types, response times, and temporal patterns.



In this study, we conduct a comprehensive comparative analysis of various data generation methods to evaluate their efficacy in producing realistic, heterogeneous multi-dimensional data suitable for simulating firefighting scenarios. We examine Random Sampling, Tabular Variational Autoencoder \cite{VAE1}, Standard Generative Adversarial Networks \cite{GAN} \cite{GAN1}, Conditional Tabular Generative Adversarial Networks \cite{TabGAN} \cite{CTGAN1}, Standard Diffusion Probabilistic Models \cite{Diff}\cite{Diff6} and Tabular Diffusion Probabilistic Models \cite{TabDDPM}.
We have chosen the methods which are the most commonly used for generating data in an industrial context. We assess their ability to capture the intricate details of firefighter interventions through a combination of standard and novel domain-specific metrics. 



Our evaluation addresses several key aspects. 
Firstly, we aim to ensure that the simulated data closely resemble the real data while avoiding exact duplication. Secondly, we introduce sufficient variability to enhance model robustness without deviating from statistical reality. Thirdly, we maintain observed biases, such as the concentration of incidents in Toulouse (France), and preserve fine correlations, like the higher frequency of chimney fires in winter or wildfires in summer. Additionally, we ensure flexibility, allowing for the testing of various scenarios in terms of frequency, type of incident, and location. We also respect and preserve the statistical anomalies observed in the real data. Lastly, we enable the generation of a large number of interventions as required to meet scalability demands.




In the following section, we present the models used in our study. The next section discusses the data and industrial context, as well as our contribution. Then, we detail the obtained results, and finally we conclude the paper and provide future perspectives.

\section{Related Work}


\textbf{Variational Autoencoders} \cite{VAE,VAE1,VAE2} are generative models that learn to encode data into a latent space and then decode from this latent space to generate new data. The model consists of two main components: an encoder and a decoder, along with a probabilistic component to enforce a structure on the latent space. The encoder maps the input data to latent variables that describes the parameters of a prior distribution (typically a Gaussian one). The decoder maps vectors sampled from the distribution described by the latent variables back to the data space. This mapping is also probabilistic since it is based on sampling in the latent space.
To enforce structure on the latent space, VAEs apply a prior distribution over the latent variables, usually a standard normal distribution, which is enforced through a Kullback-Leibler (KL) divergence regularisation loss. The reconstruction loss measures how accurately the decoder reconstructs the data from the latent variables. 


\textbf{Generative Adversarial Networks} \cite{GAN,GAN1}, consist of two neural networks: a generator \( G \) and a discriminator \( D \), which are trained simultaneously through adversarial training. The goal of the generator is to generate data that is indistinguishable from real data, while the goal of the discriminator is to distinguish between real and generated (fake) data. The generator and discriminator are trained iteratively, with each network attempting to improve against the other. The generator aims to produce data that is indistinguishable from real data, while the discriminator aims to become better at distinguishing real data from fake data. This adversarial process continues until the generator produces data that the discriminator cannot reliably distinguish from real data.



GANs are mostly used for generating images, but they have been adapted to tabular data with both nominal and numerical features in \cite{TabGAN, CTGAN1}.\\

\textbf{Diffusion models} \cite{Diff,Diff1,Diff2,Diff4,Diff5,Diff6} are a class of likelihood-based generative models that learn the data distribution using two Markovian processes. The first process, known as the forward process, progressively adds noise to the data samples, eventually transforming their distribution into an isotropic Gaussian. The second process, called the backward process, denoises the corrupted samples, allowing the generation of new samples from the learned distribution.


Adapting diffusion models for categorical data using a multinomial diffusion process was proposed in \cite{MDP}. By defining a new forward process for multinomial diffusion,  diffusion models can effectively handle both continuous and categorical data, making them versatile for various data types. Tabular Denoising Diffusion Probabilistic Model (TabDiff \cite{TabDDPM}), is designed for generating synthetic tabular data for both classification and regression tasks. This model leverages the principles of diffusion models to handle mixed-type data (both continuous and categorical).


\section{French Firefighter Context}


The Departmental Fire and Rescue Service of the French department Haute-Garonne (SDIS 31) has provided us with their intervention history for a whole year.
We remind that in our case, the data do not conform to predefined statistical models. This lack of a priori statistical patterns adds complexity to the data generation process, necessitating advanced methods that can capture the inherent irregularities and nuances. Our goal is therefore to simulate interventions that align with the spatial and temporal distribution of interventions in that particular year. Therefore, we need to generate x and y coordinates within the Haute-Garonne department, ensure a proper distribution of these interventions across the months of the year, days of the week, and hours of the day, along with the corresponding intervention duration and their variability. 


\subsection{Dataset}

The dataset is derived from the intervention data of the firefighters of Haute-Garonne for a whole year. It consists of 53,467 interventions, defined by 7 variables:

\begin{itemize}
    \item 2 continuous geographical variables: the \( x \) and \( y \) coordinates of the intervention location
    \item 3 discrete temporal variables: the hour, day, and month
    \item 1 continuous quantitative variable: the duration
    \item 1 discrete qualitative variable: the type of incident
\end{itemize}

None of the variables follow a Gaussian distribution. 

Overall, there is little correlation between the incidents and the other variables, except for certain incidents, such as chimney fires, which are almost non-existent in summer and mainly occur in winter.

\subsection{Objectives}

The goal is to create an intervention simulator to perform various tasks such as optimizing personnel and equipment resources, conducting stress tests, and anticipating different incidents. The simulated interventions need to closely resemble the provided data without being exact copies. There must be enough variability to make the model robust by training it on a multitude of possible scenarios, but this variability should not be excessive to ensure that the simulated interventions remain statistically realistic. The observed biases and fine correlations should be preserved. 

The intervention simulator should also have flexibility to test different intervention scenarios in terms of frequency, type of incident, and location. Statistical anomalies should be respected. The simulator should allow for the generation of as many interventions as needed.

\subsection{Contributions}

We compare 2 baselines, shuffling with replacement and random sampling, and 3 kinds of data generation models: 1 TVAE (a tabular variational autoencoder) \cite{CTGAN}, 2 GAN models : a standard GAN \cite{GAN} and a conditional tabular GAN \cite{CTGAN} and 2 diffusion models : a standard diffusion model \cite{Diff} TinyDiff and a tabular diffusion model TabDiff \cite{TabDDPM}. The experiment is based on a real dataset with unusual characteristics (see Dataset subsection). We evaluate these models by examining the following aspects:

\begin{itemize}
     \item We want to ensure the simulated data closely resembles the real data. We will use global distribution metrics such as Wasserstein Distance and Maximum Mean Discrepancy to measure the distance between the original distribution and the distribution of the sampled data for each model.
     \item We will ensure we introduce sufficient variability so the model trained and the sampled data do not overfit. For that, we will analyze the percentages of minimum and maximum variation for all features.
     \item Maintaining the observed biases (e.g. Preponderance of interventions in the Toulouse metropolitan area), fine correlations (e.g. chimney fires mostly in winter) and statistical anomalies (e.g. maximal duration of an intervention) is also of crucial importance from a domain expert perspective and will be accordingly assessed.
     
\end{itemize}


We demonstrate that the diffusion models, and particularly TabDiff, can be successfully used in a real-world case for data generation to support a decision-making model with an industrial application. TabDiff is effective because it associates a target variable with explanatory variables in the context of regression or classification. Although this is not our specific case, we obtained good results by designating one of our variables as a target variable. This expands the application scope of TabDiff to datasets where covariates do not follow a statistical model and are not designed for prediction.

We achieve the desired variability as well as the preservation of outliers and fine correlations and evaluate these models according to the following metrics.



\section{Results}

\subsection{Oversampling strategy for geographical constraint}

As previously discussed, achieving a realistic geographical distribution of interventions is crucial for the simulation. The interventions must be evenly distributed across predefined areas. Unfortunately, existing methods fail to meet this requirement, resulting in some areas being overloaded while others remain deserted. To address this issue, we propose an oversampling approach. We set the number of interventions in each area to the average number of interventions $\pm 2\%$. We then sample interventions using the generator until these numbers are met. Any extra interventions are discarded, and we stop when the number of generated interventions is three times (160401) the required number (53467). Under these conditions, TVAE, GAN, and CTGAN are unable to complete the generation process (with respectively only 43644, 49258 and 48964 samples). However, TabDiff and TinyDiff successfully generate the required number of interventions  with a total amount of sampled interventions of 133828 and 145792, respectively. In the following, we systematically use the oversampling approach except when indicated otherwise.

\subsection{Global distributions evaluation}
We first evaluate the results of the data generation globally to ensure that the observed statistical distributions are correctly reproduced. For this, we use four different metrics: Wasserstein Distance \cite{Wasserstein}, Maximum Mean Discrepancy (MMD) \cite{MMD}, Density and Coverage \cite{PRDC}. \\

\textbf{Wasserstein distance}, also known as Earth Mover's Distance, measures the cost of transforming one distribution into another. It takes into account the entire shape of the distributions and is sensitive to differences in their distributions, making it a robust measure for comparing real and fake data. This metric can handle cases where the support of the distributions (the range of values they can take) differs, making it versatile for various types of data. \\

\textbf{Maximum Mean Discrepancy (MMD)} is a non-parametric measure of the distance between two probability distributions. MMD does not assume any specific form of the distributions. Given two distributions \( P \) and \( Q \), MMD measures the distance between their means in a reproducing kernel Hilbert space (RKHS). The MMD between \( P \) and \( Q \) with respect to a kernel \( k \) is defined as: \[ \text{MMD}(P, Q; k) = \left\| \mathbb{E}_P[\phi(X)] - \mathbb{E}_Q[\phi(Y)] \right\|_{\mathcal{H}}.\] Here, \( \phi \) is the feature map associated with the kernel \( k \), and \( \mathcal{H} \) is the corresponding RKHS. Intuitively, MMD measures the distance between the means of the transformed samples in the high-dimensional feature space. \\

\textbf{Density} addresses the issue of overestimation that occurs with the precision metric around real data outliers. By considering the density of real samples, it provides a more accurate representation of how well the fake data aligns with the real data. Density takes into account how many real-sample neighborhood spheres contain each fake data point. This helps in understanding the local structure and distribution of the data, rather than making a binary decision based on proximity to any single real data point. Density rewards fake samples that fall within densely packed regions of real samples. This helps in ensuring that the fake data mimics the real data more closely in regions where the real data is concentrated, enhancing the overall quality of the generated data. \\

\textbf{Coverage} intuitively measures diversity by calculating the ratio of real samples that are "covered" by fake samples. This provides a direct and meaningful assessment of how well the fake data represents the variety in the real data. Coverage improves upon the recall metric by constructing nearest neighbor manifolds around the real samples rather than the fake ones. This approach is more robust because real samples typically have fewer outliers, leading to a more accurate and stable measurement. Coverage measures the fraction of real samples whose neighborhoods include at least one fake sample. This provides a clear and straightforward way to understand how well the fake data captures the real data distribution. The results are presented in Table \ref{global_metrics}.
\\

\begin{table}
\caption{Global distribution evaluations for different generative approaches and metrics. "Orig" and "Over" indicate direct sampling and oversampling, respectively.}
\resizebox{0.48\textwidth}{!}{
\begin{tabular}{llllllll}
\toprule
                & Shuffle & Rand. & TVAE     & GAN      & CTGAN    & TabDiff  & TinyDiff \\ \midrule

Precision & \textbf{0.997}  & \multicolumn{1}{r||}{0.994}  & \textbf{0.990}  & 0.985  & 0.986  & 0.988  & 0.986  \\
Recall    & 0.990  & \multicolumn{1}{r||}{0.995}  & 0.884  & 0.966  & 0.995  & 0.994  & 0.994  \\
Density   & 19.774 & \multicolumn{1}{r||}{19.519} & \textbf{22.060} & 13.018 & 18.413 & 18.744 & 17.980 \\
Coverage  & 0.989  & \multicolumn{1}{r||}{0.991}  & 0.780  & 0.912  & 0.940  & \textbf{0.973}  & \textbf{0.974}  \\               

W dist Over &         & \multicolumn{1}{r||}{   \textbf{0.049}}  & 0.506    & \textbf{0.329}    & 0.426    & 0.348    & 0.410    \\
W dist Orig         &         & \multicolumn{1}{r||}{   \textbf{0.049}}  & 0.511    & \textbf{0.340}    & 0.410    & 0.349    & 0.403    \\
MMD Over    &         &   \multicolumn{1}{r||}{   }     & 0.0474 & 0.0626 & 0.0034 & 0.0005 & \textbf{0.0001} \\
MMD Orig    &         &   \multicolumn{1}{r||}{   }       & 0.0402 & 0.0622 & 0.0020 & \textbf{0.0001} & 0.0018
\end{tabular}
}

\label{global_metrics}
\end{table}

The highest density is achieved by the TVAE model, but this comes at the cost of a significant degradation in the Coverage metric. The TabDiff model achieves the second highest density, followed by CTGAN. For coverage, the two diffusion models obtain the best scores. Regarding the Wasserstein distance, the scores obtained with or without oversampling are similar. The GAN achieves the best scores on this metric, which is expected since it directly minimizes a distribution metric loss (Kullback-Leibler distance). Finally, for the Maximum Mean Discrepancy, the two diffusion models achieve the best scores. At this level, the results seem acceptable for most of the approaches; however, some differences appear when we analyze the marginal distributions of the generated data.

\subsection{Marginal distributions evaluation}
We then evaluate the statistical distributions for each variable based on classic metrics: mean, standard deviation, minimum, and maximum values (Table \ref{stats_marginals}). We also measure the discrepancy between the distributions of categorical variables using the Jensen-Shannon divergence (Table \ref{jensen_tempo}), which measures the symmetric divergence between two probability distributions. Finally, we analyze the percentage variations to see how well the 2\% variability constraint is respected according to the different models (Table \ref{percentage}) .

\begin{table}
\caption{Statistics of marginal distributions for different generative methods. RAW is the original dataset. The closer the results are to the original values, the better the model is.}

\resizebox{0.47\textwidth}{!}{
\begin{tabular}{lrrrrrrr}
\toprule
{} &  Coord X &  Coord Y &  Month &  Day &  Hour &  Duration &  Incident \\
\midrule
Mean     &   &   &      &   &   &      &   \\ \hdashline
RAW      & 566973 & 6271183 & 6 & 179 & 13 & 88  & 11  \\
TVAE     & 567763 & 6270765 & 6 & 173 & 15 & \textbf{86}  & 1  \\
GAN      & 566494 & 6270755 & 6 & 184 & 13 & 80  & 12 \\
CTGAN    & 569352 & 6274723 & 6 & 161 & 12 & 101 & 11 \\
TabDiff  & \textbf{566939} & 6271091 & 6 & \textbf{180} & 13 & \textbf{90}  & 11 \\
TinyDiff & 566871 & \textbf{6271213} & 6 & 175 & 13 & 93  & 11 \\ \hdashline
Std     &   &   &      &   &   &      &   \\ \hdashline
RAW      & 19635 & 20112 & 3 & 103 & 6 & 58  & 13 \\
TVAE     & 16339 & 14942 & 3 & 100 & 4 & 33  & 0  \\
GAN      & 18500 & 19896 & 3 & 103 & 6 & 44  & 13 \\
CTGAN    & 14358 & 15821 & 3 & 103 & 6 & 109 & 13 \\
TabDiff  & 19493 & 20063 & 3 & 103 & 6 & \textbf{59}  & 13 \\
TinyDiff & \textbf{19539} & \textbf{20070} & 3 & 103 & 6 & 66  & 13 \\ \hdashline
Min    &   &   &      &   &   &      &   \\ \hdashline
RAW      & 492349 & 6183028 & 1 & 0 & 0 & 11 & 1 \\
TVAE     & 495807 & 6219683 & 1 & 0 & 0 & 11 & 1 \\
GAN      & 491519 & 6179851 & 1 & 0 & 0 & 11 & 1 \\
CTGAN    & 492429 & 6183093 & 1 & 0 & 0 & 11 & 1 \\
TabDiff  & 493271 & \textbf{6183028} & 1 & 0 & 0 & 11 & 1 \\
TinyDiff & \textbf{492349} & \textbf{6183028} & 1 & 0 & 0 & 11 & 1 \\ \hdashline
Max    &   &   &      &   &   &      &   \\ \hdashline
RAW      & 624065 & 6312665 & 12 & 364 & 23 & 1184 & 58 \\
TVAE     & 611449 & 6309129 & 12 & 364 & 23 & 210  & 12 \\
GAN      & \textbf{623540} & 6317500 & 12 & 364 & 23 & 274  & 58 \\
CTGAN    & 623469 & 6312661 & 12 & 364 & 23 & 1179 & 58 \\
TabDiff  & 623310 & \textbf{6312665} & 12 & 364 & 23 & \textbf{1184} & 58 \\
TinyDiff & 622062 & \textbf{6312665} & 12 & 364 & 23 & \textbf{1184} & 58 \\
\bottomrule
\end{tabular}
}

\label{stats_marginals}
\end{table}

The TVAE model is the least effective because it is biased towards the mode of the original distributions. For incidents, with a mean of 1 and a standard deviation of 0, and a maximum value of 12, it only generates 12 types of incidents out of the 58 possible types, and very often generates incident type 1. For durations, hours and coordinates, the standard deviation is significantly lower than expected, making the model unable to exhibit the desired variability. This visually results in a mass of points localized in a very small area.

The GAN performs adequately for months, days, and hours but fails to generate high values for intervention durations. Moreover, the minimum and maximum values of the coordinates indicate that the model generates interventions located outside the Haute-Garonne department.


The CTGAN performs better than the GAN on maximum durations but its standard deviation of 109 is nearly double that of the original distribution. Additionally, the standard deviations of the coordinates are too low, meaning that the interventions are located in the same place. It is noteworthy that the GAN performs well on days.


\begin{table}[]
\caption{Jensen-Shannon divergence for temporal features. }

\resizebox{0.47\textwidth}{!}{
\begin{tabular}{lllllll}
\toprule
{}  & Shuffle & TVAE & GAN & CTGAN & TabDiff & TinyDiff \\
\midrule \hdashline

JSD  &   & &  &  &  &  \\\hdashline
Month    &   \multicolumn{1}{r||}{\textbf{0.269}} &  6.150 & 3.014 & 4.529 &   \textbf{0.734} &   \textbf{1.926} \\
Day     &   \multicolumn{1}{r||}{\textbf{0.339}} &  2.019 & 0.878 & 2.093 &   \textbf{0.864} &    \textbf{0.810} \\
Hour    &    \multicolumn{1}{r||}{\textbf{0.344}} & 17.486 & 3.266 & 7.425 &   \textbf{1.087} &    \textbf{1.128} \\
Incident &   \multicolumn{1}{r||}{\textbf{ 0.355}} & 17.733 & 2.627 & 0.973 &   \textbf{0.353} &    \textbf{0.525} \\
Area &  \multicolumn{1}{r||}{\textbf{0.334}} & 24.022 & 3.885 & 4.159 &   \textbf{0.161} &    \textbf{0.161} \\

\bottomrule
\end{tabular}
}

\label{jensen_tempo}
\end{table}

The two diffusion models achieve the best scores. The TabDiff model performs slightly better on months. 

\begin{table}[]
\caption{Percentage of minimum and maximum variation for all features.}

\resizebox{0.47\textwidth}{!}{
\begin{tabular}{lrrrrr}
\toprule
{} &  TVAE &  GAN &  CTGAN &  TabDiff &  TinyDiff \\
\midrule
\% Min Month        &      \textbf{ 1} &      7 &       \textbf{ 1 } &      \textbf{    1 } &        \textbf{   1 } \\
\% Max Month        &      55 &     89 &       43 &        \textbf{ 13 } &      \textbf{    13 } \\ \hdashline
\% Min Day        &     \textbf{  0 } &    \textbf{  0 } &        1 &       \textbf{   0 } &         \textbf{  0 } \\
\% Max Day        &     145 &    113 &      213 &         69 &        \textbf{  59 }\\ \hdashline
\% Min Hour       &       5 &      2 &     \textbf{   0 } &      \textbf{    0 } &           2 \\
\% Max Hour       &     210 &     80 &      128 &       \textbf{  12 } &          22 \\ \hdashline
\% Min Duration       &      48 &    \textbf{  0 } &      \textbf{  0 } &       \textbf{   0 } &        \textbf{   0 } \\
\% Max Duration       &    \textbf{  98  } &    676 &      700 &        300 &         700 \\ \hdashline
\% Min Incident    &      91 &    \textbf{  0  }&        1 &       \textbf{   0  } &         \textbf{  0  } \\
\% Max Incident    &     249 &    356 &       26 &     \textbf{    22  }&          73 \\ \hdashline
\% Min Area &    \textbf{   0   } &  \textbf{    0  }&       \textbf{ 0} &       \textbf{   0 } &       \textbf{    0 } \\
\% Max Area &     100 &     60 &       90 &        \textbf{  2 } &         \textbf{  2 } \\
\bottomrule
\end{tabular}
}

\label{percentage}
\end{table}

The only two models that respect the 2\% variability constraint are the diffusion models. These models perform the best, with overall fairly low variations across different variables. Only the TVAE performs better on the variation of the maximum duration, with only 98\%. However, this is because the model generates a very low maximum duration.

\subsection{Domain expert evaluation}
\begin{figure}
    \centering
    \includegraphics[width=\linewidth]{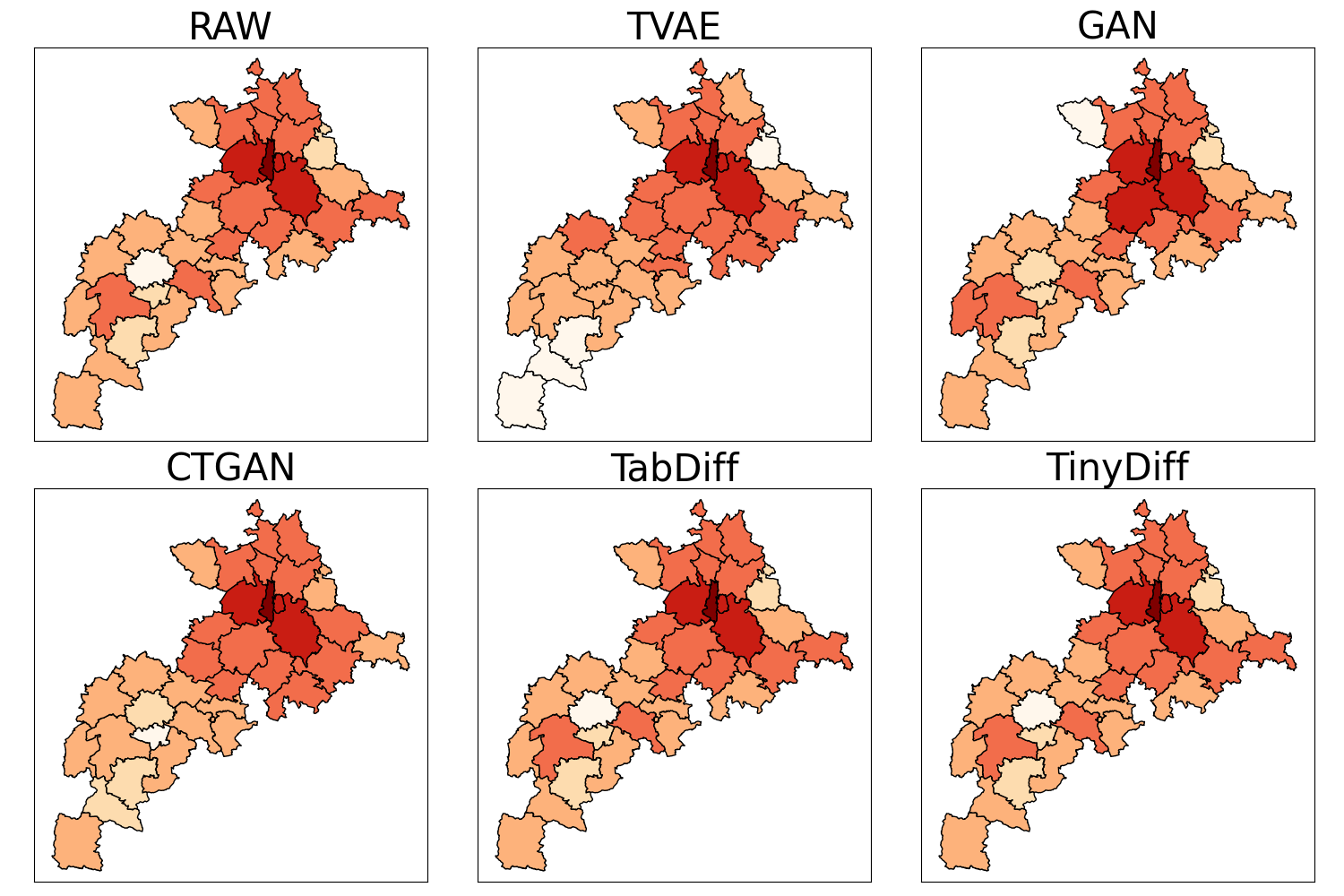}
    \caption{number of interventions by area, RAW is the original dataset}
    \label{fig:areas}
\end{figure}

The only models that faithfully represent the number of interventions per area are the diffusion models (Figure \ref{fig:areas}). The other models show too much variability or fail to generate interventions in certain areas, as is the case with the TVAE, making them unsuitable for heterogeneous tabular data synthesizing.

\begin{figure}
    \centering
    \includegraphics[width=\linewidth]{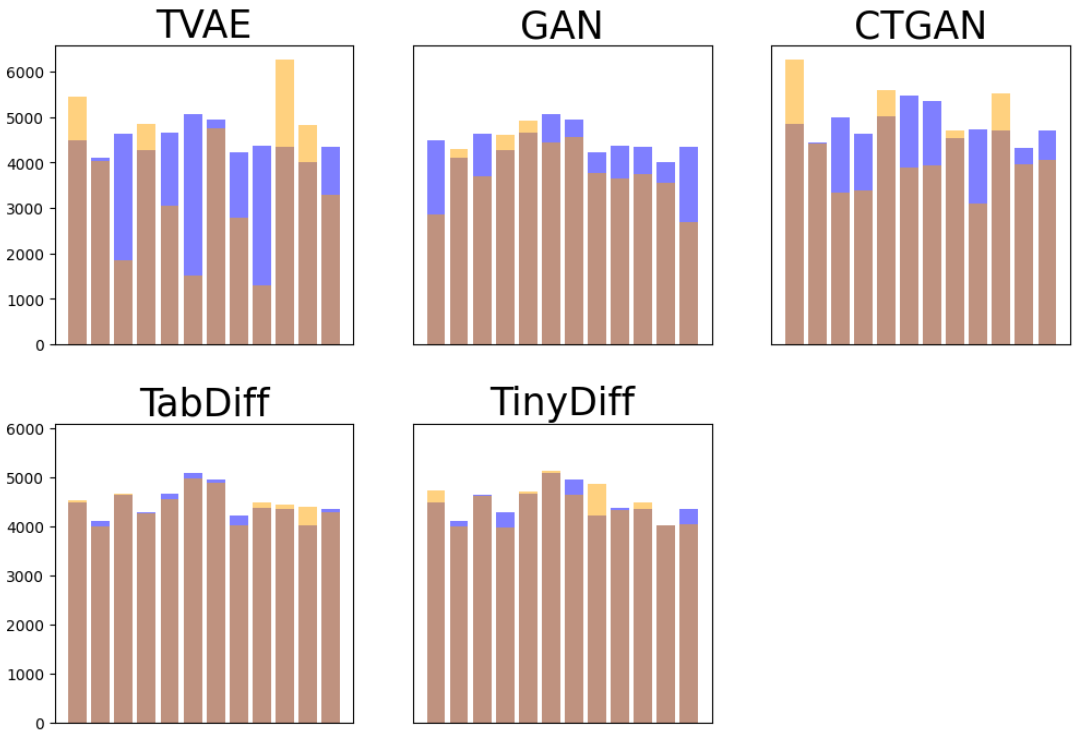}
    \caption{number of interventions by month (blue : true data, orange : simulated data) }
\label{fig:inter_by_month}
\end{figure}

\begin{figure}
    \centering
    \includegraphics[width=\linewidth]{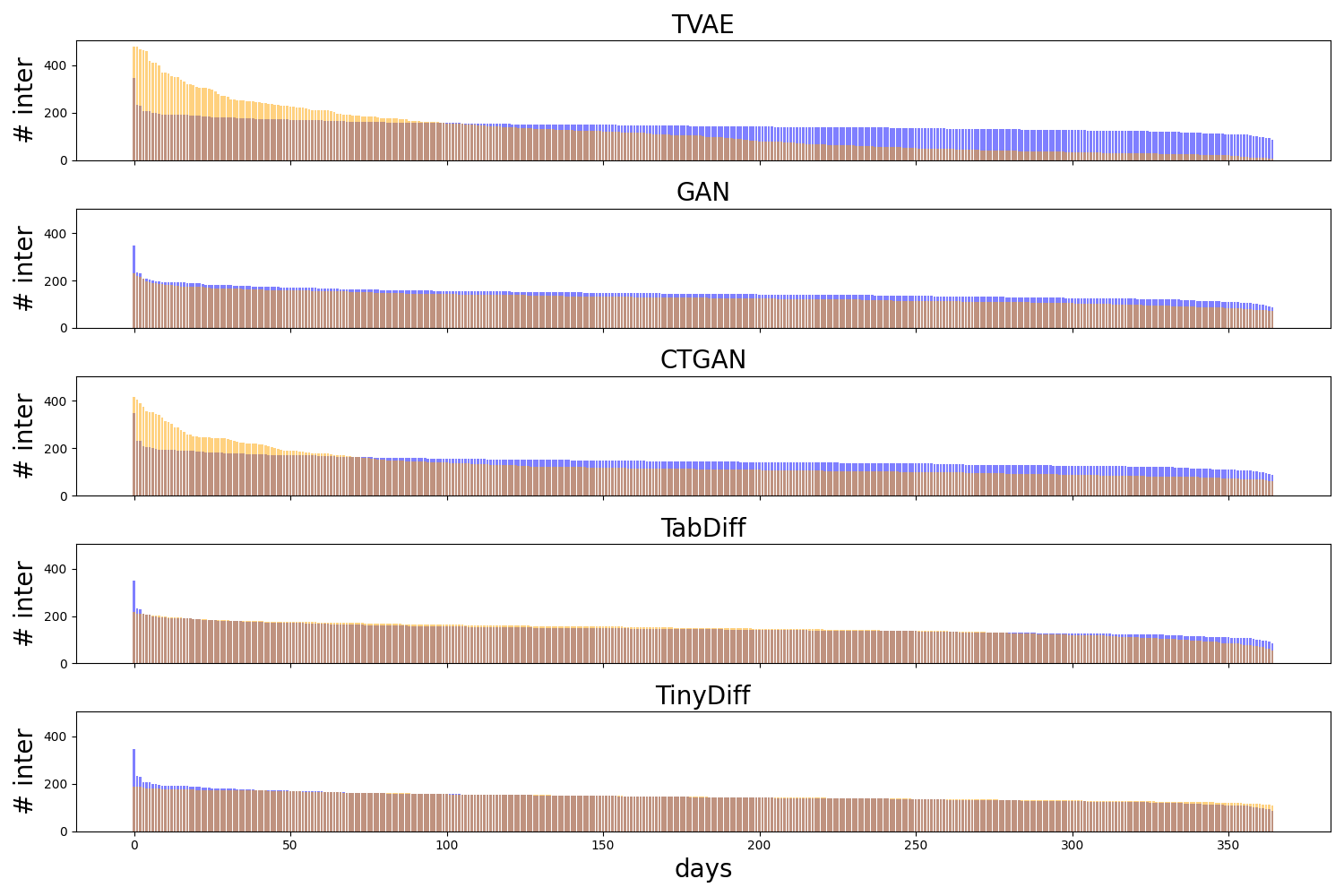}
    \caption{number of interventions by day, sorted by descending order (blue : true data, orange : simulated data) }
\label{fig:inter_by_day}
\end{figure}

These results are also reflected in the number of interventions per day and per month (Figures  \ref{fig:inter_by_month} and \ref{fig:inter_by_day}). The diffusion models capture general trends while ensuring low variability. The other models exhibit significant imbalances depending on the month or day.

\begin{figure}
    \centering
    \includegraphics[width=\linewidth]{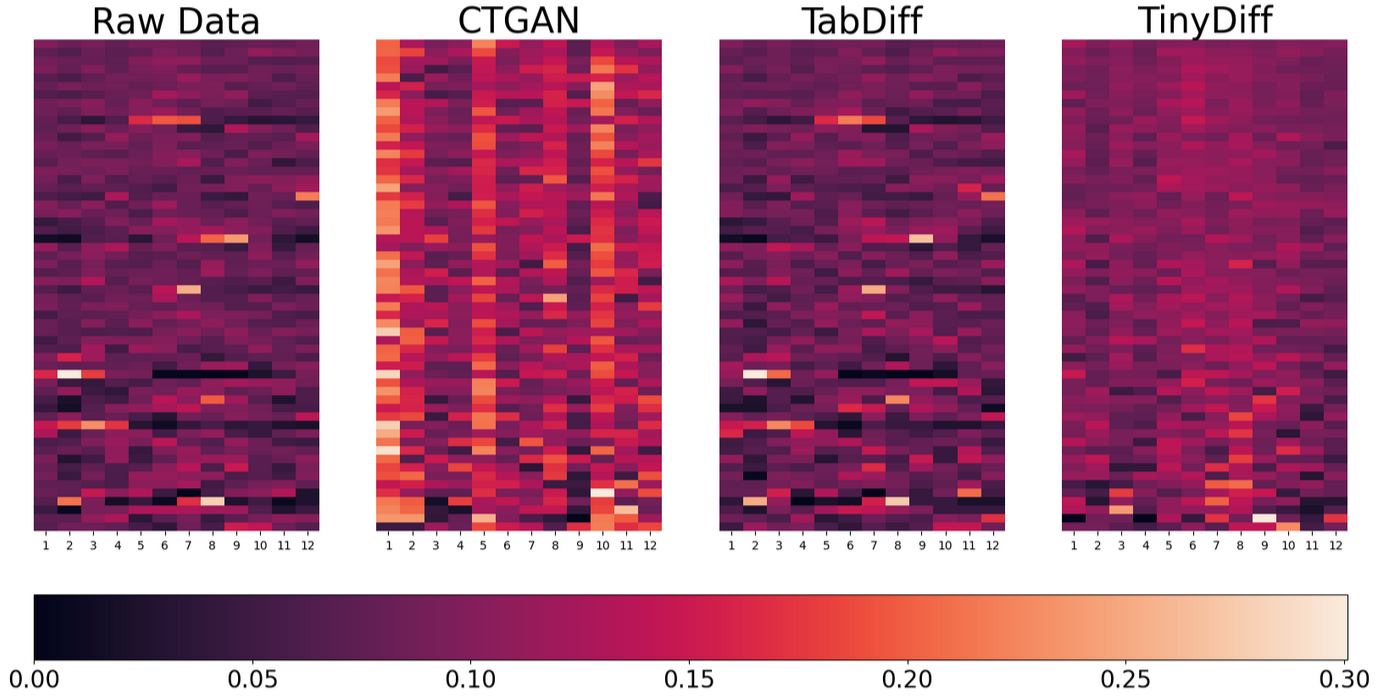}
    \caption{Correlations between types of incident and months}
    \label{fig:correlations}
\end{figure}

Figure \ref{fig:correlations} shows the observed correlations between the occurrences of different types of incidents on the y-axis and the months on the x-axis for four datasets: from left to right, the original data, those generated by CTGAN, by TabDiff, and by TinyDiff. The lighter the color, the stronger the correlation. Only the TabDiff model correctly captures the fine correlations that exist between the occurrence of certain types of incidents and specific months. This is because it considers the incident variable as a variable to predict, unlike the TinyDiff model, which is therefore unable to generate interventions that conform to the observed correlations between incident types and months.

\subsection{Simulation of interventions}
To ensure that the generated data is useful for our application, we conducted an intervention simulation using both the real dataset and the dataset generated by the TabDiff model. This simulation models the logic allocating personnel to vehicles and dispatching vehicles based on the type of incident and the location of the intervention, using the real archived data of firefighters, vehicles, and fire stations available in Haute-Garonne. This corresponds to 53,467 interventions over one year for each dataset. We then tracked the following metrics: the number of vehicles currently involved in interventions across the entire department at the start of each intervention, and the number of vehicles deployed for interventions throughout the year for each vehicle type.

\begin{figure}
    \centering
    \includegraphics[width=\linewidth]{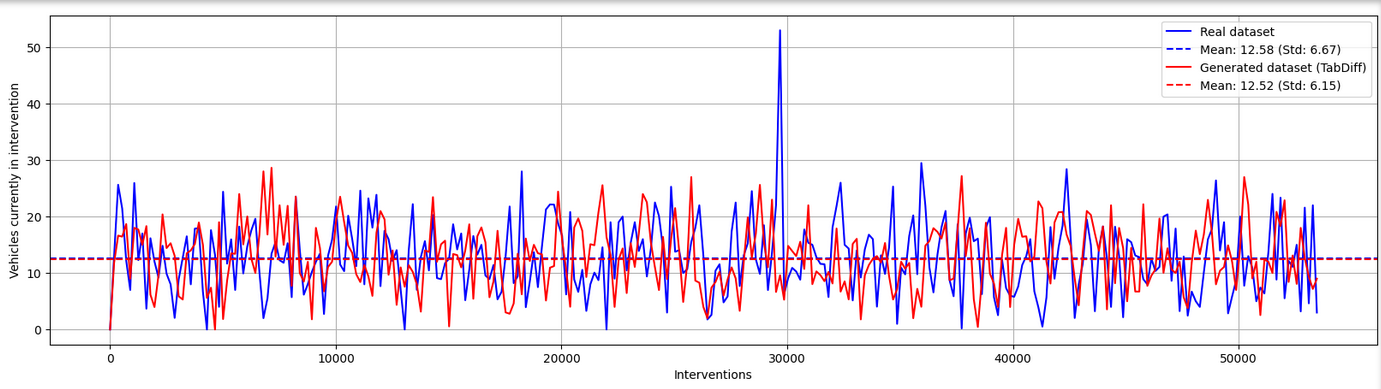}
    \caption{Mean of vehicles in intervention}
    \label{fig:mean_out}
\end{figure}

In Figure \ref{fig:mean_out}, we can see that the total average number of vehicles in intervention is nearly identical (12.58 vs. 12.52), and the standard deviations are also similar (6.67 vs. 6.15). The slight difference observed can be attributed to a day in the real dataset when a major flood occurred, leading to the deployment of numerous vehicles. However, as this event is extremely rare in the dataset, the data generation model did not manage to reproduce such a disaster with the same impact. This is not an issue, as it is a rare (unique) event and, by definition, unpredictable, and it is preferable to integrate such events manually if necessary.

\begin{figure}
    \centering
    \includegraphics[width=\linewidth]{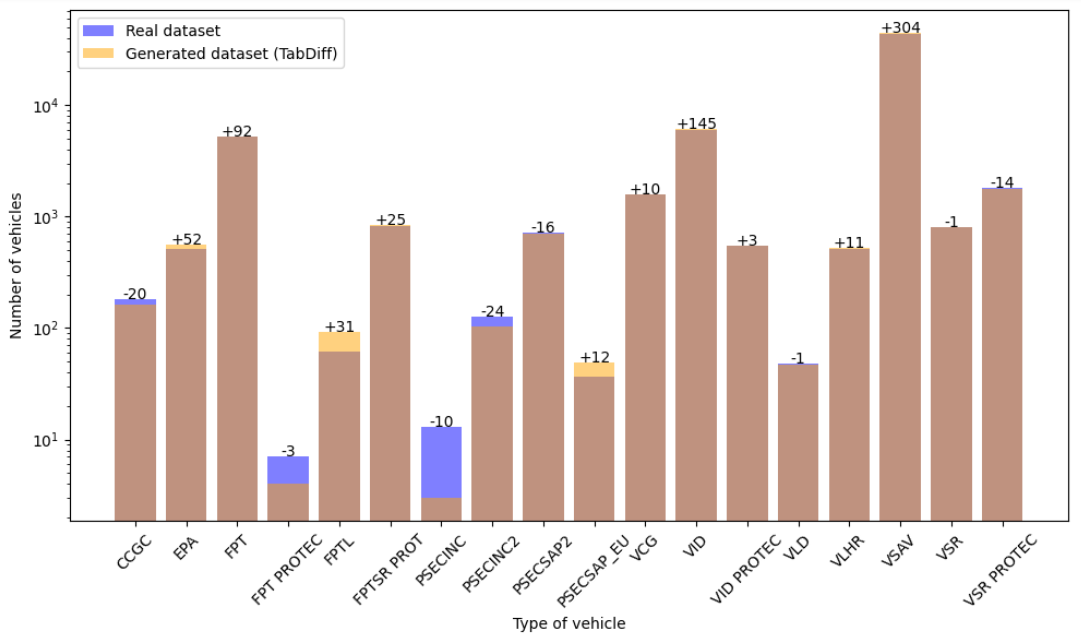}
    \caption{Total of vehicles sent in intervention, by type}
    \label{fig:type_v_int}
\end{figure}

In Figure \ref{fig:type_v_int}, we observe that the number of vehicles mobilized during the year is almost identical for each type of vehicle. It is important to note that the y-axis is on a logarithmic scale. Thus, the variation of 304 observed for the vehicle type "VSAV" should be considered in relation to the total number of VSAVs mobilized, which is approximately 44,000 per year. Similarly, the variation observed for the "PSECINC" vehicle type should be put into perspective given the low frequency of interventions involving this vehicle. Overall, it is evident that the number of vehicles deployed is very similar between the real dataset and the generated dataset, and the observed variations are within the normal range of statistical variation for a sample, even though the types of interventions and their chronological order are completely different between the two datasets.

This demonstrates that our generated data can be used for practical applications beneficial to firefighters, such as intervention simulations.

\section{Conclusion}

Our findings revealed significant disparities in the performance of the models. TVAE, GAN, and CTGAN struggled to complete the generation process, producing substantially fewer samples than required. In contrast, TabDiff and TinyDiff succeeded in generating the necessary number of interventions, demonstrating their robustness and scalability. The global evaluation, using metrics like  Maximum Mean Discrepancy (MMD) and Coverage, indicated that diffusion models (TabDiff and TinyDiff) outperformed the other methods. These evaluation metrics make more sense from an application point of view than Wasserstein or Density, which favor GAN or TVAE respectively.  They maintained the observed statistical distributions and respected the variability constraints, producing realistic and diverse synthetic data.  In the marginal evaluation, diffusion models again showed superior performance, accurately capturing the distributions of individual variables and adhering to the 2\% variability constraint. While TVAE and GAN exhibited significant limitations in generating realistic data, CTGAN performed better but still fell short compared to the diffusion models. The final assessment using precision and recall metrics further emphasized the diffusion models' advantage. Despite TVAE achieving the highest density, it compromised coverage, whereas TabDiff and TinyDiff balanced both metrics effectively. The GAN model, though excelling in Wasserstein distance due to its loss function, generated fewer samples than required, underscoring the limitations of traditional GANs in this context.


Only diffusion models can capture the various observed trends, whether it's the frequency of interventions per day or per month, or the distribution of interventions by sector. The TabDiff model, which is the only one that properly handles the incident variable, is also able to maintain the fine correlations between the types of incidents and the months. 

It is also the only model that can generate data matching the observed patterns and trends, making it suitable for practical tasks such as simulating interventions for operational planning and analysis.  







\bibliography{AAAI2025/aaai25}
\bibliographystyle{plain}

{\footnotesize \textbf{Code}: \url{https://github.com/TheSachari/Data_Generation} \\
\textbf{Datasets}: \url{https://www.dropbox.com/scl/fo/vz49vv8tsbg5fquy690dp/ALUDJ2F49mSXhqzsddP_xF0?rlkey=sxs7lf2xlbgd8ndx3ctpqgztc&st=icwfdi5g&dl=0}
}

\end{document}